\documentclass[]{llncs}
\usepackage{graphicx}

\usepackage{tikz}
\usepackage{comment}
\usepackage{amsmath,amssymb} 
\usepackage{color}

\usepackage[accsupp]{axessibility}  

\usepackage{amsmath,amsfonts,bm}

\def\eqref#1{equation~\ref{#1}}

\def\1{\bm{1}}

\newcommand{\test}{\mathcal{D_{\mathrm{test}}}}

\DeclareMathAlphabet{\mathsfit}{\encodingdefault}{\sfdefault}{m}{sl}
\SetMathAlphabet{\mathsfit}{bold}{\encodingdefault}{\sfdefault}{bx}{n}

\DeclareMathOperator*{\argmax}{arg\,max}

\usepackage[pagebackref,breaklinks,colorlinks]{hyperref}
\usepackage[utf8]{inputenc} 
\usepackage[T1]{fontenc}    
\usepackage{url}            
\usepackage{booktabs}       
\usepackage{amsfonts}       
\usepackage{nicefrac}       
\usepackage{microtype}      
\usepackage{xcolor}         
\usepackage{multirow}
\usepackage{pifont}
\newcommand{\cmark}{\ding{51}}%
\newcommand{\xmark}{\ding{55}}%
\usepackage{amsmath}
\usepackage{graphicx}
\usepackage{caption}
\usepackage{subcaption}

\begin{document}
\pagestyle{headings}
\mainmatter
\def\ECCVSubNumber{5238}  

\title{ A Simple Approach to Adversarial Robustness in Few-shot Image Classification  }
\titlerunning{ECCV-22 submission ID \ECCVSubNumber}
\authorrunning{ECCV-22 submission ID \ECCVSubNumber}
\author{Anonymous ECCV submission}


\titlerunning{}
%
\author{Akshayvarun Subramanya\inst{1} \and
Hamed Pirsiavash\inst{2}}
\authorrunning{}
%
\institute{University of Maryland, Baltimore County \and
University of California, Davis}
\maketitle

\newcommand{\fix}{\marginpar{FIX}}
\newcommand{\new}{\marginpar{NEW}}

\begin{abstract}

  Few-shot image classification, where the goal is to generalize to tasks with limited labeled data, has seen great progress over the years. However, the classifiers are vulnerable to adversarial examples, posing a question regarding their generalization capabilities. Recent works have tried to combine meta-learning approaches with adversarial training to improve the robustness of few-shot classifiers. We show that a simple transfer-learning based approach can be used to train adversarially robust few-shot classifiers.  We also present a method for novel classification task based on calibrating the centroid of the few-shot category towards the base classes.  We show that standard adversarial training on base categories along with calibrated  centroid-based classifier in the novel categories, outperforms or is on-par with state-of-the-art advanced methods on standard benchmarks for few-shot learning. Our method is simple, easy to scale, and with little effort can lead to robust few-shot classifiers. Code is available here: \url{https://github.com/UCDvision/Simple_few_shot.git}
\end{abstract}

\section{Introduction}
Few-shot learning presents the challenge of generalizing to unseen tasks with limited data. The problem is aimed at learning quickly from few examples of data, which is generally considered the hallmark of human intelligence.
This is an important practical problem due to the scarce availability of fully annotated data in the real world. Researchers have shown that such a setting can be considered for various real world computer vision tasks such as image classification \cite{finn2017model,chen2019closer}, object detection \cite{wang2020tracking}, image segmentation \cite{rakelly2018conditional}, face-recognition \cite{Guo_2020_cvpr} and medical analysis \cite{maicas2018training}. As a result, it is of paramount importance that such safety-critical systems are reliable and robust to changes in input. Specifically in this work, we consider robustness to adversarial examples - carefully crafted perturbations using gradients that when added to inputs, \textit{fool} the classifier. The most common method of improving robustness is by adversarial training \cite{43405} which involves training on adversarial examples using adversary of choice. Traditional adversarially robust methods \cite{madry2018towards,43405,intriguing-arxiv-2013} consider a data-rich setting where many examples are available per category. It has also been shown that adversarial generalization  possibly requires significantly more data \cite{schmidt2018adversarially}. This becomes challenging in a scenario where the end-user has access to limited amount of annotated data  but is interested in building a robust few-shot classifier. Such a setting is more practical and it is important to develop methods which can work with minimal effort in the pre-deployment stage. We show from our experiments that a simple approach of finetuning the network on clean data from an adversarially robust base model can lead to significant improvement in robustness with minimal resources.\\
Previous works on improving robustness for few-shot classifiers focus mainly on meta-learning approaches where the base model is trained on adversarial examples of episodic data. We show that standard mini-batch based adversarial training is sufficient for learning a robust classifier.  This makes our method simple and scalable, making the process of training robust classifiers straightforward and also creating directions to explore methods from robustness literature.
\vspace{-0.05in}

{\noindent}It is important to understand the problem of few-shot learning in order to develop their robust counterparts. The goal in few-shot learning is to learn transferable knowledge for generalization to tasks with limited data.
These have generally been partitioned into metric learning \cite{snell2017prototypical,vinyals2016matching,sung2018learning},  optimization-based \cite{finn2017model,ravi2016optimization} and hallucination based methods \cite{hariharan2017low,yang2021free,antoniou2017data,wang2018low}. The most common work among optimization based methods is MAML \cite{finn2017model} which aims at learning a network initialization using a bi-level optimization procedure, that when finetuned on limited data is able to generalize to the new task. Recent works have shown that meta-learning methods can be extended to include adversarial robustness as well. \cite{goldblum2019robust,wang2021fast} perform adversarial training on top of meta-learners to improve robustness significantly. However, adversarial training on its own is expensive and combining with meta-learning makes the problem computationally intensive. \cite{wang2021fast} showed that there exists a compromise between training robust meta-learners and performance, motivating the need for a simpler approach.
\vspace{-0.1in}

\noindent Another interesting line of work, focused on improving few-shot learning, shows that a simple method which involves training on large scale data and finetuning the model on the few-shot dataset can match or even outperform meta-learning methods \cite{chen2019closer,dhillon2019baseline}. The intuition is that the model sees examples from all categories and can get a general sense of semantics rather than seeing only episodic data. Such a setting also makes it easier to train on large scale data that can lead to further improvements as shown in \cite{dhillon2019baseline}. Our results show considering a simple setting can be beneficial for adversarial robustness as well.\\
\noindent  We consider the few-shot setting and show that adversarial training along with a centroid-based classifier can outperform previous methods in terms of robustness. Such a setting is practically relevant, since the adversarial training is done just once and robustness for few-shot classes can be achieved without creating adversarial examples. We believe it also becomes easier to incorporate new approaches to robustness, such as verifiably robust classifiers \cite{cohen2019certified,gowal2018effectiveness} and can bring together robust methods for both large and limited dataset settings.

 \noindent In the following sections, we describe our method and discuss relevant related work. We present experimental findings and implementation details in the subsequent sections followed by conclusion and directions for future work.

\begin{figure}[!h]
    \centering
    \subfloat[\centering Robust Base Training ]{{\includegraphics[width=0.6\textwidth]{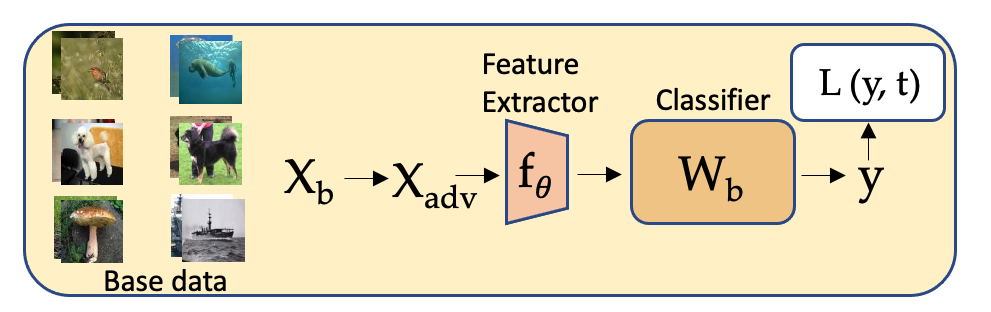} }}%
    \qquad
    \subfloat[\centering Novel Training]{{\includegraphics[width=0.6\textwidth]{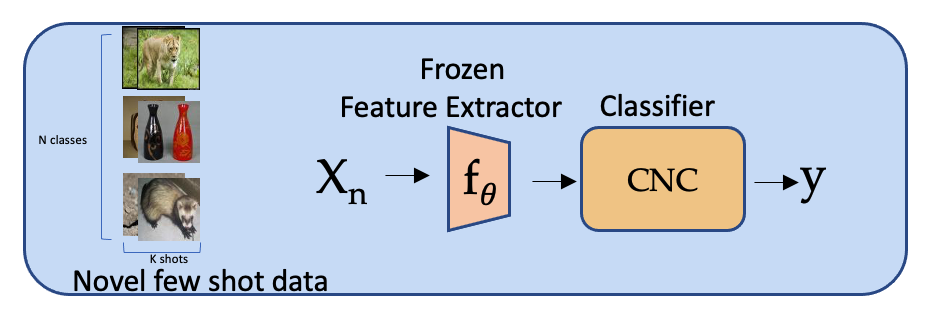} }}%
    \caption{Our method is divided into two phases (a) Robust Base training involves standard adversarial training with base dataset consisting of many examples and categories (b) Novel Training involves transferring or adapting the network for novel few-shot data using a Calibrated Nearest Centroid (CNC) classifier. Note that the feature extractor remains frozen in the second phase.}%
    \vspace{-0.1in}
    \label{fig:example}%
\end{figure}

\section{Method}
Here we introduce notation and provide a description of our method. Our first objective is to learn a feature extractor $f_{\theta_b}$ and linear classifier $C_{\omega_b}$ using the abundantly-labeled base dataset $X_b$. Previous approaches consider multiple few-shot tasks sampled from $X_b$ for meta-learning, whereas we consider a standard mini-batch based training.\\
At the next stage, when a $N$-way $K$-shot few-shot task is sampled from the novel dataset $X_n$, we  use only the feature extractor $f_{\theta_b}$ and learn a new linear classifier  $C_{\omega_n}$ such that it can generalize to unseen examples from novel categories. We show that a simple approach of training a robust base model and then adapting it to novel categories can outperform previous approaches.
We divide our approach into two stages: (1) Robust Base training and (2) Novel training.
\subsection{Robust Base Training}
\label{sec:rob_base}
Given a base dataset $X_b$ with large number of annotated examples per category, we perform adversarial training using an iterative adversary such as PGD \cite{madry2018towards}. Specifically, we solve the min-max objective
\begin{equation}
\theta^{*}=\min_\theta \mathbb{E}_{(x,y) \in X_b}\bigg[\max_{|| \delta ||_p<\epsilon} \mathcal{L} (\theta,x+\delta,y)  \bigg]
\label{eqn:adv}
\end{equation}

Here, $\mathcal{L} (\theta,x,y)$ represents the training objective, which is commonly cross-entropy and $\theta = (\theta_b,\omega_b)$ represents the combination of the feature extractor and base classifier parameters. There are different methods for optimizing the inner maximization in Equation \ref{eqn:adv}. We use the Projected Gradient Descent (PGD) algorithm, an iterative algorithm presented in \cite{madry2018towards} with $p=\infty$ which corresponds to finding a perturbation $\delta$ around an $\epsilon$-bounded hypercube around $x$ that maximizes the objective. Once we find the perturbation, the perturbed input is added to the training set and parameters are tuned. This method is called adversarial training and is the most widely used method to improve robustness to adversarial examples. Note that adversarial training, which is a computationally expensive procedure, needs to be performed just once using base dataset. The intuition behind this approach is that the model sees multiple categories across batches rather than episodic data, hence gaining better understanding of semantic categories and robustness which can be beneficial for adaptation.



\noindent \textbf{Weight averaging:} Weight averaging (WA) has been shown as a simple way to improve generalization \cite{izmailov2018averaging,garipov2018loss} in deep networks as it approximates ensembling in temporal fashion and can find flatter optima in loss surface. This method has been used in adversarial training \cite{gowal2020uncovering,chen2021robust} for improving robustness in standard classification task.  Since we are interested in using base parameters at the next stage, we perform weight averaging for only the feature extractor parameters $\theta_b$ and show this can be used for few-shot setting.

\noindent Similar to \cite{gowal2020uncovering}, we keep a separate copy of the weights and for every iteration perform exponential moving average method $\theta{_b}' \leftarrow \tau \theta{_b}' + (1- \tau) * \theta{_b} $ and use $\theta{_b}'$ during the evaluation. We set $\tau=0.999$ in all our experiments.
\vspace{-0.1in}
\subsection{Novel Training}
During this stage, we consider the $N$-way, $K$-shot method as the novel task and adapt our learnt feature extractor $f_{\theta_b}$ using classifier $C_{\omega_n}$ . For all our experiments, we found best results when the weights of feature extractor $f_{\theta_b}$ are frozen and not optimized during novel training. Intuitively, this can be understood not wanting the parameters of the feature extractor to be biased towards the few-shot examples. And since we are interested in learning only from clean data during the novel training, there can also be an effect of \textit{forgetting} the robustness learnt at the base stage. This was observed in \cite{goldblum2019robust} where only the final layer was trained and rest of the parameters were frozen. During novel training, we use only clean examples and not adversarial examples, making the process straightforward.

\noindent \textbf{Linear classifier:} The simplest possible baseline is to learn a linear model on top of the frozen feature extractor using the few shot examples of novel categories. As shown in our experiments, this simple baseline on its own achieves reasonable performance compared to previous approaches. This baseline also suggests that a robust base classifier corresponds to  a robust novel classifier and a simple approach such as ours is enough to  achieve robustness for few-shot classifiers. However, as observed in previous works \cite{wang2021fast,goldblum2019robust}  and in our experiments, this approach alone is not sufficient to achieve improved robustness. Interestingly, we achieve much closer results to state of the art compared to previous works using this simple baseline. One challenge associated with using few-shot data is that the model can become biased towards the specific samples and may not capture the true class distribution. Hence there is a need for more calibrated classifiers.



\noindent \textbf{Background on Distribution Calibation (DC) \cite{yang2021free}:} Recent work \cite{yang2021free}  has shown that standard accuracy of few-shot classifiers can be improved by using Distribution Calibration. They present a \textit{free-lunch} hallucination-based method where the feature distributions of the novel categories are calibrated using the base dataset, due to the similarity between the base and novel datasets. The mean and covariance of each novel category is calibrated using the statistic of base data. They use these statistics to \textit{hallucinate} or  sample many points from a Gaussian distribution, and learn a logistic regression classifier.  This method was shown to improve standard accuracy significantly under various settings.

\noindent \textbf{Calibrated Nearest Centroid (CNC):}
DC method can be computationally expensive due to the calculation of covariance matrix which can be of $\mathcal{O}(N*D^2)$ complexity where $D$ is the dimensionality of the feature space and $N$ is the number of data points in the base dataset. The covariance matrix is also expensive to store in memory. Moreover, sampling from a multivariate Gaussian with non-diagonal covariance is also expensive and can be of the order of at least $\mathcal{O}(D^{2.3})$  \cite{bishop:2006:PRML}. These can reduce the applicability of the approach for large scale settings.

\noindent Another aspect of using the hallucinated features is that the model can become biased to the clean features and learn a non-robust final classifier. Note that since these additional data points are generated in the feature space and not the image space, it is not possible to create adversarial versions of these features and perform adversarial training using the large set of features. This poses a problem of improving the performance without sacrificing robustness.

\noindent To overcome these drawbacks, we present a simple method where we rely only on the \textbf{calibrated mean} and classify query sample using a non-parametric Nearest Centroid based algorithm. We call this the \textbf{Calibrated Nearest Centroid} (CNC) Classifier. We find the nearest base-category centers to each novel training sample and then average them along with the novel training sample to obtain the new mean or centroid for the novel category. We do not consider the covariance matrix in our method and we find this approximation works equally well in our experiments. More formally:
\begin{equation}
    \mu_j = \frac{1}{m+1} (z_j+\sum_{i \in \mathcal{S}_j} \mu^b_i)
\end{equation}

\noindent where $\mu_j$ is the center for the novel category $j$ and $\mathcal{S}_j$ is the set of $m$ base category centers that are closest to $z_j$,~~~$\mu^b_i$ is the mean of base category $i$ in the feature space. In the case of $k$-shot setting, we calculate a centroid for each sample and average them to get one centroid for each category.


\noindent At inference time, we simply find the nearest center to the query point and assign its label:
\begin{equation}
    \hat y= \{ y_j | \argmax_j \tilde{\mu}_j^T \tilde{z} \}
    \label{eqn:cosine}
\end{equation}
\begin{figure}[t]
    \centering
    \subfloat[\centering Nearest Centroid Classifier without Calibration ]{{\includegraphics[width=5cm]{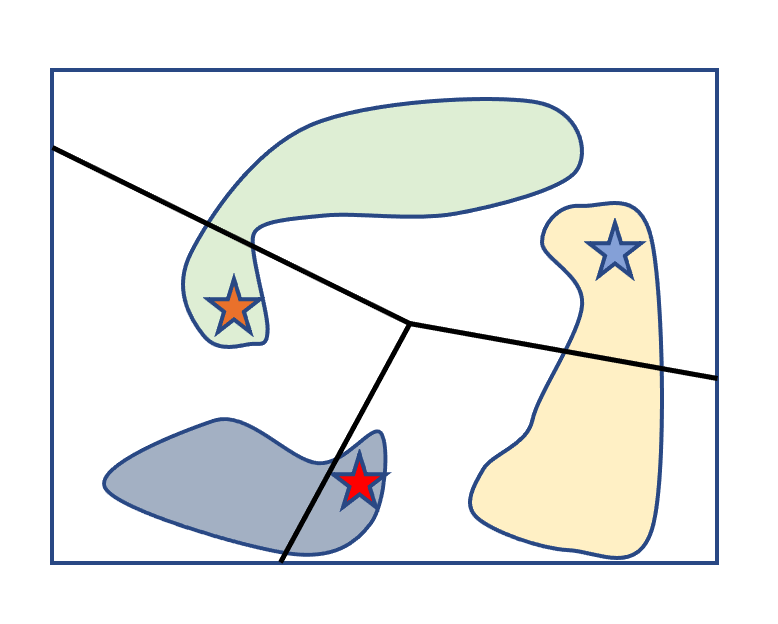} }}%
    \qquad
    \subfloat[\centering Calibrated Nearest Centroid Classifier]{{\includegraphics[width=5cm]{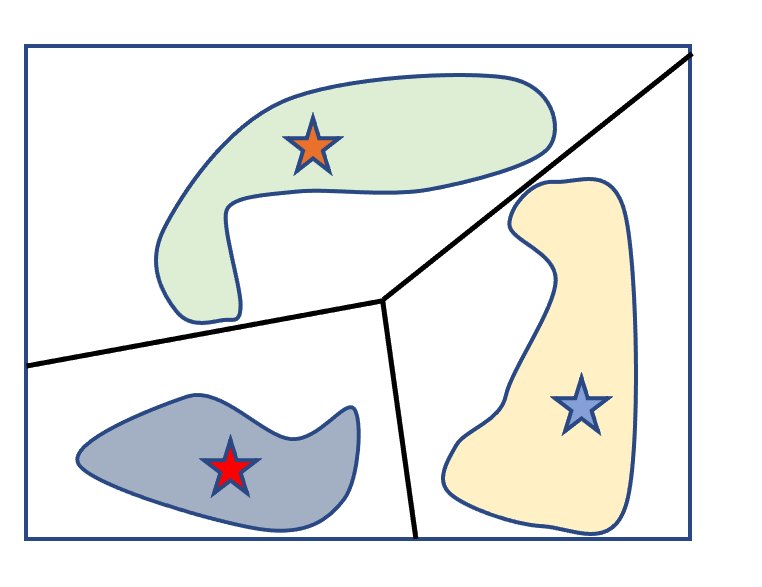} }}%
    \caption{An illustration of the calibration for the Nearest Centroid Classifier. Since we are working in the few-shot regime from novel dataset, if the examples are not sampled from across the distribution, it may result in centroids which do not necessarily represent the true distribution. On the other hand, calibration using the base classes can result in centroids which better represent the actual distribution of the class, allowing for better generalization. }%
    \label{fig:example}%
\end{figure}


\noindent where $\tilde{z}$ and $\tilde{\mu}$ represent $\ell_2$ normalized version of vectors $z$ and $\mu$ respectively, while $\hat y$ is the predicted category. Note that $\ell_2$ normalization is performed for both query point and centroids. Since we consider the normalized version of the vectors, euclidean distance reduces to the form in Equation \ref{eqn:cosine}, similar to recent works \cite{grill2020bootstrap}. Our inference  can be considered similar to the Linear Classifier method except that our centres ( $\mu_j$ ) are estimated with simple averaging rather than learnt using SGD. Note that similar to \cite{yang2021free}, as a preprocessing step, we transform the embeddings by taking the square root of each dimension so that their distribution gets closer to Gaussian form. It can be seen as ``Tukey's Ladders of Power Transformation'' \cite{tukey1977exploratory} with $\lambda=0.5$.



 \noindent At the inference time, a Nearest Centroid-based method requires less memory and computation compared to a Nearest Neighbor classifier since we only have to store and compare with one prototype per class rather than entire training set. 



\section{Related Work}

\noindent \textbf{Few shot Image Classification:} Few-shot learning is a challenging problem in computer vision where the goal is rapid generalization to unseen tasks. Metric learning approaches such as \cite{snell2017prototypical,vinyals2016matching,sung2018learning} were some of the earliest approaches towards tacking this problem. \cite{snell2017prototypical} learn a metric space where prototypical representation of each category is utilized for classifying novel data. \cite{ravichandran2019few} showed that rather than taking the average, learning the prototypes along with the model can lead to better performance.
More recently, a family of algorithms based on learning to learn \cite{andrychowicz2016learning} or meta-learning have gained considerable attention. \cite{ravi2016optimization} develop an LSTM based meta-learner to train anothet network for the few-shot task.  \cite{finn2017model,nichol2018first}  create a model agnostic algorithm which aims to learn a good initialization that can adapt to new tasks. Hallucination based methods such as \cite{hariharan2017low,wang2018low,antoniou2017data} also present promising directions towards improved generalization. \cite{gidaris2018dynamic,qi2018low} try to directly predict the weights of the classifier for novel categories. \cite{yang2021free} calibrate the distributions of few shot examples using the statistics of categories with larger number of examples. Our calibration is similar to \cite{yang2021free}, but we apply it at category-level rather than instance-level. However recent works have shown that simple baselines which are non-episodic in nature can provide competitive performance for few-shot image classification task \cite{chen2019closer,dhillon2019baseline}. Such line of work provide for non-sophisticated baselines and pose a question to the community to rethink the approach towards few shot learning.

\noindent \textbf{Adversarial examples:} Adversarial examples are carefully crafted perturbations designed to fool the model \cite{intriguing-arxiv-2013,43405,world-arxiv-2016}. \cite{43405} showed that adversarial examples can be created rather easily using the sign of single gradient step, which they called as Fast Gradient Sign Method (FGSM). The existence of such examples raises a question regarding the generalization capabilities of Deep Neural networks. Many defenses have been proposed to overcome this problem \cite{papernot2016distillation,xie2017adversarial,feinman2017detecting,filterstats-arxiv-2016}, but they have been bypassed with slight modifications to the adversary \cite{athalye2018obfuscated,carlini2016defensive,carlini2017adversarial}. One of the most common approaches called adversarial training involves incorporating the adversarial examples into the training set. \cite{madry2018towards} showed that the first order adversary, based on the Projected Gradient Descent (PGD) algorithm can be used to train robust neural networks. Provable methods have developed which can provide certification on the susceptibility of an input towards adversaries \cite{wong2018provable,wong2018scaling,raghunathan2018semidefinite,gowal2018effectiveness}. A class of algorithms based on randomized smoothing \cite{cohen2019certified,yang2020randomized} have shown promising results in training large scale neural networks. \cite{zhang2019theoretically} provide a theoretical analysis on the robustness vs accuracy trade-off, which had been studied empirically \cite{tsipras2018robustness}, and show that their algorithm named TRADES improves robustness compared to previous approaches. \cite{shafahi2019adversarial} use the gradients from the backpropagation algorithm to improve robustness with minimal cost.

\noindent \textbf{Adversarial Robustness for Few-shot classifiers:} Recent works have tried to address the problem of adversarial examples in the context of few-shot learning. \cite{yin2018adversarial} used the FGSM adversary to create adversarial examples and optimized a meta-learner to be robust to adversarial examples. \cite{goldblum2019robust} showed that meta-learning algorithms can be supplemented with adversarial examples in the query set to learn robustness. Their method called Adversarial Querying was shown to be robust to strong attacks such as PGD. Many meta-learning approaches were extended to their robust counterparts by including adversarial query examples. Recently, \cite{wang2021fast} also proposed a similar approach where MAML was used as the base meta learning algorithm. \cite{wang2021fast} also showed that including a contrastive learning objective similar to \cite{chen2020simple} can provide a way to use unlabelled data when learning the model and thus improve both standard and robust accuracy.
\vspace{-0.15in}
\section{Experiments}
\vspace{-0.15in}
In this section, we describe our experiments and provide implementation details.
We consider three benchmark datasets - \textbf{Mini-ImageNet}, \textbf{CIFAR-FS} and \textbf{CUB} for our experiments. \textbf{Mini-ImageNet} consists of 100 classes derived from ImageNet dataset \cite{ILSVRC15} where each category consists of 600 images. This was first proposed in \cite{vinyals2016matching} and recent works follow the split provided by \cite{ravi2016optimization} consisting of 64 base, 16 validation and 20 novel classes. We use images of size  84x84 for all experiments with Mini-ImageNet. \textbf{CIFAR-FS} was proposed in \cite{bertinetto2016learning} as a benchmark for few-shot classification. It splits CIFAR-100 dataset similar to Mini-ImageNet. \textbf{CUB} \cite{wah2011caltech} is a fine-grained dataset which has been used as a benchmark for few-shot classification. It consists of 11788 images split across 200 categories. We use the split provided by \cite{hilliard2018few} consisting of 100 base , 50 validation and 50 novel classes. As per our knowledge, we are the first to show adversarial robustness for a fine-grained dataset under few-shot setting. Note that there is no intersection of categories across base and novel subsets for all datasets.
We use Pytorch \cite{paszke2017automatic} framework and NVIDIA 2080Ti GPUs for our experiments.

 \noindent {\bf Implementation details:} \label{sec:impl}  In the base training stage, as described in Section \ref{sec:rob_base}, we follow the attack parameters of \cite{goldblum2019robust} and use iterative PGD attack with 7 iterations during training with $\epsilon=8/255$ and $\alpha=2/255$ for all experiments unless otherwise mentioned. We mainly use the standard network architecture ResNet \cite{He_2016_CVPR} and also provide results for other architectures. For weight averaging, we set the parameter $\tau=0.999$. We use SGD  optimizer with a learning rate of 0.1 and weight decay of $1e-5$ for the feature extractor parameters $\theta_b$ and $1e-4$ for the classifier parameters $\omega_b$. We train the model for 250 epochs with a batch size of 64.
 For novel training and learning the Linear Classifier, we follow the setting described in \cite{chen2019closer} and learn the parameters using SGD with momentum $0.9$ and learning rate $\eta=0.01$. We set the dampening as $0.9$ and weight decay of $1e-3$. For our calibration, we use $m=2$ number of base categories.
 For calculating Robust Accuracy, we use 20 iterations of PGD to create an adversarial example and measure accuracy w.r.t ground truth. We report both Standard Accuracy and Robust Accuracy for 5-way, 1-shot and 5-way, 5-shot settings averaged over 1000 different trials on the test set. We report the mean of the different trials as well as the 95\% confidence intervals for all our experiments. Since our goal is to build robust models, we mainly focus on improving Robust Accuracy.\\

\noindent We compare our results with \cite{goldblum2019robust} ,which we refer to as Adversarial Querying (AQ) where adversarial examples are created for query data and
\cite{wang2021fast} which we refer as \textbf{OFA} where MAML \cite{finn2017model} is used to adversarially train the model. We refer to training a linear classifier during the novel training stage as \textbf{Linear} and the Calibrated Nearest Centroid classifier as \textbf{CNC}.


\subsection{Mini-ImageNet}
\vspace{-0.1in}
 We present our main results on Mini-ImageNet in Table \ref{tab:mini_main}. We observe that \textbf{CNC} method outperforms other approaches in Robust Accuracy under most settings and  boosts standard accuracy as well. We also show results on large-scale architectures such as WideResNets \cite{zagoruyko2016wide} and DenseNets \cite{huang2017densely}.  The difference becomes clear as we move to larger architectures. Our Linear classifier also serves as a strong baseline for robust few-shot settings.

 \begin{table}[!h]
    \centering
    \resizebox{0.98\linewidth}{!}{
   \begin{tabular}{|l |c||c|c | c | c| c|  }
\hline
& & \multicolumn{2}{|c|}{1-shot } & \multicolumn{2}{|c|}{5-shot }\\
\cline{3-6}
 Method & Backbone  & Standard & Robust & Standard & Robust\\
 &  & Acc. & Acc. & Acc. & Acc.\\
\hline
AQ & ResNet18 & 41.48 $\pm$ 0.56 & 20.52 $\pm$ 0.45  & 59.32 $\pm$ 0.53 & 32.18 $\pm$ 0.50  \\

Ours (Linear) &ResNet18 & 42.63  $\pm$ 0.56  & 19.56$\pm$ 0.45  & \textbf{61.35 $\pm$ 0.51} & 30.63 $\pm$ 0.52 \\
Ours (CNC) &ResNet18 & \textbf{44.98 $\pm$ 0.59} &\textbf{ 21.38 $\pm$ 0.46}  & \textbf{61.30 $\pm$ 0.55} & \textbf{33.41 $\pm$ 0.51} \\

\hline
AQ &  WRN-50-2 & 38.99 $\pm$ 0.55 & 22.09 $\pm$ 0.45  & 57.11 $\pm$ 0.51 & 33.62 $\pm$ 0.50  \\
Ours (Linear) &WRN-50-2& 43.14  $\pm$ 0.54   & 19.94 $\pm$ 0.43   & 62.93  $\pm$ 0.50  & 30.52 $\pm$  0.52 \\
Ours (CNC) &WRN-50-2& \textbf{46.71  $\pm$ 0.62 } & \textbf{23.04 $\pm$ 0.50}  & \textbf{63.60 $\pm$ 0.55 }& \textbf{36.06 $\pm$ 0.54}  \\
\hline
AQ & WRN-28-10 &44.17 $\pm$ 0.60 & \textbf{23.81 $\pm$ 0.48}  & 62.41 $\pm$ 0.54 & 33.62 $\pm$ 0.50 \\

Ours (Linear) &WRN-28-10& 52.36 $\pm$ 0.62    &  22.23$\pm$0.52 & \textbf{72.11 $\pm$ 0.51} & 32.29  $\pm$ 0.59  \\

Ours (CNC) &WRN-28-10& \textbf{53.22 $\pm$ 0.66} & 22.91 $\pm$ 0.51    & 70.13 $\pm$  0.52 & \textbf{35.40 $\pm$ 0.58  } \\

\hline
AQ & DenseNet121 & 38.32 $\pm$ 0.55  & 10.19 $\pm$ 0.32  & 56.65 $\pm$ 0.51 & 17.77 $\pm$ 0.41  \\

Ours (Linear) &DenseNet121& 39.77 $\pm$  0.56  & 18.16$\pm$0.42& 57.45$\pm$ 0.54  & 27.89 $\pm$0.52  \\

Ours (CNC) &DenseNet121& \textbf{ 42.05$\pm$ 0.60 } &\textbf{ 20.21 $\pm$ 0.45 }  & \textbf{58.59$\pm$0.55} & \textbf{ 32.24$\pm$0.56 } \\

\hline

AQ &DenseNet161 & 37.35 $\pm$ 0.52  & 9.80 $\pm$ 0.31   & 55.97  $\pm$ 0.53 & 16.69  $\pm$  0.38 \\

Ours (Linear) &DenseNet161& 40.75 $\pm$  0.55  & 17.44 $\pm$ 0.41& 59.84 $\pm$ 0.53  &  27.11$\pm$ 0.50 \\

Ours (CNC) &DenseNet161& \textbf{43.48 $\pm$ 0.60}  & \textbf{20.63 $\pm$ 0.45}   & \textbf{60.92 $\pm$ 0.54} & \textbf{33.87 $\pm$ 0.53} \\

\hline
\end{tabular}}
    \caption{\textbf{Results on Mini-ImageNet dataset}. Our CNC method outperforms other approaches which becomes clear as we move to larger architectures. We can also see that our linear classifier serves as a strong baseline and can be used to learn robust few-shot classifier.}
    \vspace{-0.4in}
    \label{tab:mini_main}
\end{table}

 \noindent Our method is straightforward to scale for large architectures since they are equivalent to standard adversarial training. However, we observed that scaling meta-learning combined with PGD adversarial training is a difficult task. \\
 To provide a comparison, we trained both AQ and our model on 4 NVIDIA TITAN RTX GPUs using WideResNet-28-10 backbones. AQ method took 1.7 hour/epoch and required 60 epochs while our method took 0.36 hour/epoch for 250 epochs. The total training time for AQ was around 100 hours whereas our method took around 90 hours. This shows that our approach is more easily scalable compared to previous methods.


 \begin{table}[!h]
    \centering
   \begin{tabular}{|l|c ||c|c | c | c|   }
\hline
& &  \multicolumn{2}{|c|}{Conv4 } & \multicolumn{2}{|c|}{ResNet18}\\
\cline{3-6}
 Method & Base training &  Standard & Robust & Standard & Robust\\
 & & Acc. & Acc. & Acc. & Acc.\\
\hline
AQ & AT & 29.6 & 24.9   & 30.04 & 20.05  \\
OFA & AT &\textbf{40.82} & 23.04 &  38.94 & 19.94 \\
OFA &TRADES & 37.1  & 25.51   & 41.94 & 20.19  \\
OFA &CL &  38.60 &  26.81  & 43.98 & 21.47 \\
Ours (Linear)& AT & 38.39 $\pm$ 0.37  & 28.76 $\pm$ 0.33 & 44.93$\pm$ 0.37 & 29.30 $\pm$ 0.33  \\
Ours (CNC) & AT& 39.23 $\pm$ 0.38 & \textbf{30.77 $\pm$ 0.35}  & \textbf{49.15 $\pm$ 0.41} &  \textbf{35.59  $\pm$ 0.38}   \\
\hline

\hline
\end{tabular}
    \caption{Comparison with \cite{wang2021fast} on Mini-ImageNet dataset. Note that both TRADES and CL use additional unlabelled data in a semi-supervised manner. Our method outperforms previous approaches on both settings.
    }
    \vspace{-0.3in}
    \label{tab:ofa_mini}
\end{table}

 {\noindent \bf Comparison with OFA:} We compare with another recent work OFA \cite{wang2021fast} where MAML was combined with adversarial training to improve robustness. We present them separately compared to previous results as the attack parameters and testing configuration followed are different. For a fair comparison, we use the same setting and we refer the readers to \cite{wang2021fast}.  \textbf{Base training} column indicates the type of adversary used during base dataset training where AT indicates PGD adversarial training, TRADES is the algorithm presented in \cite{zhang2019theoretically} and CL corresponds to using the Contrastive Learning objective \cite{chen2020simple}. Both TRADES and CL use additional unlabelled data in a semi-supervised manner. We observe from Table \ref{tab:ofa_mini} that our method has clear gains in terms of robust accuracy and surpasses standard accuracy in some cases as well.
 \subsection{CIFAR-FS}

Here we present results on CIFAR-FS dataset. The organization is similar to Mini-ImageNet dataset where we presented two tables for fair comparison with previous approaches.
To compare with Adversarial Querying (AQ) \cite{goldblum2019robust}, we use a ResNet18 backbone and use attack parameters $\epsilon=8/255$ and $\alpha=2/255$ with 20-iteration PGD for testing.
As seen in Table \ref{tab:cifar_1_main}, our Calibrated Nearest Centroid classifier outperforms previous approaches on Robust Accuracy, which is the main focus of our work. We also observe that we match or outperform on Standard Accuracy under both 1-shot and 5-shot settings.

\begin{table}[!t]
    \centering
    \resizebox{0.98\linewidth}{!}{
   \begin{tabular}{|l |c||c|c | c | c| c|  }
\hline
& & \multicolumn{2}{|c|}{1-shot } & \multicolumn{2}{|c|}{5-shot }\\
\cline{3-6}
 Method & Backbone  & Standard & Robust & Standard & Robust\\
 &  & Acc. & Acc. & Acc. & Acc.\\
\hline
AQ & ResNet18 &  45.41$\pm$ 0.68 & 21.76  $\pm$  0.59  & \textbf{64.98 $\pm$ 0.58} & 34.24  $\pm$ 0.65  \\

Ours (Linear) &ResNet18& 44.76 $\pm$ 0.63  & 21.01  $\pm$ 0.58 &  62.23 $\pm$  0.63  &  31.60$\pm$ 0.66  \\

Ours (CNC) &ResNet18& \textbf{ 48.89$\pm$ 0.71  } & \textbf{ 27.16 $\pm$ 0.66 } & \textbf{ 64.36$\pm$ 0.61 }  & \textbf{ 39.13 $\pm$ 0.71}  \\
\hline
\end{tabular}}
    \caption{\textbf{Results on CIFAR-FS dataset}. We can see our CNC method outperforms compared to previous approaches. This experiment uses the same attack parameters as \cite{goldblum2019robust}.}
    \label{tab:cifar_1_main}
    \vspace{-0.4in}
\end{table}

\noindent For fair comparison with \cite{wang2021fast}, we train a model with Conv4 backbone and use their attack parameters. The results are presented in Table \ref{tab:ofa_cifar}. As explained earlier, \textbf{Base training} column indicates the type of adversary used during base dataset training where AT indicates PGD adversarial training, TRADES uses the algorithm presented in \cite{zhang2019theoretically} and CL corresponds to using the Contrastive Learning objective \cite{chen2020simple}. We see a similar trend as before where our CNC outperforms previous approaches. We also conduct an experiment using TRADES during Base Training, allowing us to compare methods that use similar adversary. Under such comparable settings, our method outperforms previous approaches. This experiment shows that our method can generalize to other adversarial training methods and we believe that as more advanced methods are developed in the community, they can be incorporated in a straightforward manner to improve robustness under few-shot settings.

 \begin{table}[!h]
  \vspace{-0.1in}
    \centering
   \begin{tabular}{|l |c||c|c | c | c| c|  }
\hline
& & \multicolumn{2}{|c|}{1-shot } & \multicolumn{2}{|c|}{5-shot }\\
\cline{3-6}
 Method & Base training  & Standard & Robust & Standard & Robust\\
 &  & Acc. & Acc. & Acc. & Acc.\\
\hline
AQ & AT & 31.25 & 26.34  & 52.32 & 33.96  \\
OFA &AT & 39.76 & 26.15 & 57.18 & 32.62 \\
OFA &TRADES & 40.59 & 28.06  & 57.62 & 34.76 \\
OFA &CL & 41.25 & \textbf{29.33}  & \textbf{57.95} & 35.3 \\
Ours (Linear) &AT& 41.12 $\pm$ 0.40  & 25.65 $\pm$ 0.37   & 56.20  $\pm$ 0.39 & 34.73  $\pm$ 0.41 \\
Ours (CNC)& AT &  41.81$\pm$ 0.41 &  28.22 $\pm$ 0.40   & 53.52 $\pm$ 0.40  &  39.09 $\pm$ 0.42  \\
Ours (CNC)& TRADES & \textbf{43.56$\pm$0.43}  & 28.12  $\pm$ 0.41   & 56.99  $\pm$ 0.40   &  \textbf{39.48 $\pm$0.43} \\

\hline
\end{tabular}
    \caption{Comparison with \cite{wang2021fast} for Conv4 backbone on CIFAR-FS dataset. Comparing methods that use same base training procedure (\textbf{AT} or \textbf{TRADES}  ) , we can see that our CNC method outperforms on Robust Accuracy under both 1-shot and 5-shot settings. This experiment shows that our method can generalize to other adversarial training methods as well.}
    \vspace{-0.3in}
    \label{tab:ofa_cifar}
\end{table}

 \subsection{CUB}
 For results on CUB dataset, we use the same attack parameters described for Mini-ImageNet i.e, $\epsilon=8/255$, $\alpha=2/255$, 7 iterations of PGD during training and 20 during testing. We use ResNet18 backbone and implement AQ as per the guidelines given in \cite{goldblum2019robust} and our best implementation is presented in Table \ref{tab:cub_main}.

\noindent Since CUB is a fine-grained classification dataset, the base and novel categories share greater similarity compared to previous datasets. Hence it serves as an opportunity to understand how the robustness transfers from base to novel dataset, i.e whether the similarity in classes acts as a boon or bane under fine-grained dataset settings. As seen from our experiments, the linear classifier baseline performs reasonably well indicating that a robust base classifier transfers to a robust novel classifier. The CNC method also benefits under such settings and outperforms all other methods on both Standard and Robust Accuracy.
\begin{table}[!h]
    \centering
    \resizebox{0.98\linewidth}{!}{
   \begin{tabular}{|l |c||c|c | c | c| c|  }
\hline
 && \multicolumn{2}{|c|}{1-shot } & \multicolumn{2}{|c|}{5-shot }\\
\cline{3-6}
 Method & Backbone  & Standard & Robust & Standard & Robust\\
 &  & Acc. & Acc. & Acc. & Acc.\\
\hline
AQ &ResNet18 & 54.27 $\pm$ 0.79 & 28.23 $\pm$ 0.66   & 68.42 $\pm$ 0.62  & 37.10 $\pm$ 0.66  \\

Ours (Linear) &ResNet18& 51.93 $\pm$ 0.71   &  27.24 $\pm$ 0.64  & 69.83  $\pm$ 0.61  & 37.06  $\pm$ 0.68 \\

Ours (CNC) &ResNet18& \textbf{56.42 $\pm$ 0.78  } & \textbf{32.18$\pm$ 0.70} & \textbf{71.51  $\pm$ 0.60}  & \textbf{44.33 $\pm$ 0.69}  \\
\hline
\end{tabular}}
    \caption{\textbf{Results on CUB dataset}. We show that robustness transfers from base to novel datasets under fine-grained classification setting as well. Our Linear classifier serves as a strong baseline and our CNC method outperforms on both metrics. }
    \label{tab:cub_main}
    \vspace{-0.5in}
\end{table}
 \subsection{TieredImageNet}
We also consider the large scale dataset TieredImageNet\cite{ren2018meta}, which is a subset sampled  hierarchically from ILSVRC12. We use the standard split of 351 base classes and 160 novel classes. Each class is a child of one of the 34 more abstract categories from ImageNet. Hence, the classes are spread differently compared to MiniImageNet. This allows us to test our method against varying hardness and diversity of the few-shot categories. The similarity between the base and novel categories is also varied in this context and shows that our method performs well under such settings. As seen in Table \ref{tab:tier_1_main} that the linear classifier acts as a strong baseline and our CNC method outperforms previous works.

\begin{table}[!h]
    \centering
   \begin{tabular}{|l |c||c|c | c | c| c|  }
\hline
& & \multicolumn{2}{|c|}{1-shot } & \multicolumn{2}{|c|}{5-shot }\\
\cline{3-6}
 Method & Backbone  & Standard & Robust & Standard & Robust\\
 &  & Acc. & Acc. & Acc. & Acc.\\
\hline
AQ &  ResNet18 & 49.77 $\pm$ 0.70 & 29.78  $\pm$ 0.65   & 66.72 $\pm$ 0.56  &  43.73 $\pm$ 0.63   \\

Ours (Linear) &ResNet18& 50.47  $\pm$ 0.69  & 27.90  $\pm$ 0.60 & 68.48  $\pm$  0.60  & 40.30 $\pm$ 0.65   \\

Ours (CNC) &ResNet18& \textbf{51.38 $\pm$ 0.71} & \textbf{30.27 $\pm$ 0.62} & \textbf{68.50$\pm$ 0.59}  &  \textbf{44.64$\pm$ 0.66} \\
\hline
\end{tabular}
    \caption{Results on TieredImageNet dataset. }
    \label{tab:tier_1_main}
    \vspace{-0.3in}
\end{table}

\section{Discussion}
\label{sec:discussion}
In this section, we provide additional intuition into our method and also showcase some benefits of our simple framework. We also refer the readers to appendix for more experiments and analysis.
\begin{table}[!h]
\vspace{-0.1in}
\centering
\resizebox{\linewidth}{!}{
 \begin{tabular}{|c | c || c |c | c|| c| c |c| c |}
 \hline
 & Base  & \multicolumn{3}{|c||}{Novel }& \multicolumn{2}{|c|}{1-shot } & \multicolumn{2}{|c|}{5-shot } \\
 & Training  & \multicolumn{3}{|c||}{Training}&\multicolumn{2}{|c|}{} & \multicolumn{2}{|c|}{}\\ [0.5ex]
 \hline
Exp. & WA & DC & CNC & Linear& Standard & Robust  & Standard & Robust  \\
Id &   &  & & &Acc.& Acc.  & Acc. & Acc.  \\
 \hline \hline
1& \xmark & \xmark & \xmark  &\cmark & 41.40 $\pm$ 0.56 & 18.25 $\pm$ 0.45  & \textbf{59.30 $\pm$ 0.54 }& 27.96 $\pm$ 0.50 \\
 \hline
2&  \xmark & \cmark & \xmark  &\xmark & \textbf{43.72 $\pm$ 0.57} & 14.18 $\pm$ 0.38  & 58.04 $\pm$ 0.52 & 13.97 $\pm$ 0.39 \\
 \hline
3&  \xmark & \xmark & \cmark  &\xmark & 42.56 $\pm$ 0.60 & \textbf{19.57 $\pm$ 0.45 } & 58.22 $\pm$ 0.53 & \textbf{30.42 $\pm$ 0.50} \\
 \hline \hline

4&  \cmark & \xmark & \xmark  &\cmark & 42.63 $\pm$ 0.56 & 19.56 $\pm$ 0.45   & \textbf{61.35  $\pm$ 0.51 } & 30.63 $\pm$ 0.52 \\
 \hline
5&  \cmark & \cmark & \xmark  &\xmark & 44.73 $\pm$ 0.59  & 15.29  $\pm$ 0.41   & 59.78  $\pm$ 0.53  & 19.49  $\pm$ 0.49 \\
 \hline
6& \cmark & \xmark & \cmark  &\xmark & \textbf{44.98  $\pm$ 0.59} & \textbf{21.38  $\pm$ 0.46}   & 61.30 $\pm$ 0.55  & \textbf{33.41  $\pm$ 0.51} \\

 \hline

 \hline
\end{tabular}
}
\caption{Illustration of different configurations of Base and Novel training. Here we show results on ResNet18 backbone on Mini-ImageNet dataset. WA represents Weight Averaging, DC represents Distribution Calibration and CNC corresponds to the Calibrated Nearest Centroid classifier.  }
\vspace{-0.3in}
\label{tab:mini_resnet18}
\end{table}
\\
{\noindent \bf Analysis :}  Here, we would like to study the effect of different combinations of the base and novel training, allowing us to analyse them carefully. We perform an experiment on Mini-ImageNet dataset on ResNet18 backbone similar to Table \ref{tab:mini_main} and the results are presented in Table \ref{tab:mini_resnet18}. We observe that the simple baseline of robust base training and training a linear classifier during novel training works reasonably well which can be a considered a strong baseline (Exp Id 1,4). We also observe that DC algorithm proposed in \cite{yang2021free} improves standard accuracy but introduces a drop in robustness (Exp Id 2,5). Note that DC involves hallucination of examples at a feature level and learning a Logistic Regression Classifier. We believe that one reason for the drop in robustness is because the final classifier becomes biased to the clean hallucinated examples, leading to non-robust margins across different classes. In most configurations we observe that our method matches or even outperforms the DC method and more importantly improves robustness  (Exp Id 5,6). This shows that the robustness of the base dataset is transferred to the novel setting and the calibration based Nearest Centroid classifier can improve robustness in few-shot setting.  The impact of weight averaging (WA) method can be observed when considering Exp Ids 3 and 6. The temporal ensembling nature of the method helps in finding flatter minima, thereby boosting performance. Previous works \cite{gowal2020uncovering} have shown that this can be used for standard robust classifiers. Here, we observe this holds true under a transfer-learning type setting. We find that the combination of these methods can lead to improved performance.


 {\noindent \bf Extension to verifiably robust models:} An advantage of our simple framework is that we can incorporate methods from the adversarial examples literature for few-shot learning. Specifically, we consider verifiably robust procedures where the goal is to provide a guarantee on adversarial robustness of the model. Standard adversarial training methods do not lead to provably robust models  leading to low verified accuracy, we observe a similar trend in our experiments as well. Many methods have been proposed in this area \cite{mirman2018differentiable,zhang2019towards,cohen2019certified} and we consider one of them - Interval Bound Propagation (IBP) \cite{gowal2018effectiveness}. This allows to train a provably robust model whose accuracy does not drop below the verified accuracy for a given threat model.
 Here we show that replacing PGD-training with IBP during the robust base training stage described in section \ref{sec:rob_base} can lead to verifiable robustness for few-shot classifiers. We show results for 1-shot setting in Table \ref{tab:ver_cifar} where we use a ResNet18 backbone on CIFAR-FS dataset. We use the training procedure as described in \cite{xu2020automatic} with $\epsilon=8/255$ for 1000 epochs. Here Robust Accuracy refers to 20-iteration PGD testing and Verified Accuracy is calculated similar to \cite{gowal2018effectiveness}.  We believe this experiment can encourage researchers to incorporate more advanced verification methods in the future and develop algorithms for few-shot settings.

 \begin{table}[!h]
    \centering
   \begin{tabular}{|l ||c| c|c| }
\hline
 &  \multicolumn{3}{|c|}{1-shot } \\
\cline{2-4}
  Method &   Standard & Robust &Verified\\
 &   Acc.  & Acc. & Acc.\\
\hline

IBP + Linear  & 37.01  $\pm$ 0.65   & 26.77 $\pm$ 0.59 &   21.79$\pm$ 0.55   \\

IBP + CNC &  37.72 $\pm$  0.65 & 28.12 $\pm$ 0.62 & 23.25 $\pm$ 0.61\\

\hline

\hline
\end{tabular}
    \vspace{0.1in}
    \caption{Extension to verifiably robust classifiers. We show that it is possible to train verifiably robust models for few-shot settings. This is an added advantage of our framework due to the similarity to standard classifier training. Results are shown using ResNet18 backbone on CIFAR-FS dataset. }
    \label{tab:ver_cifar}
    \vspace{-0.3in}
\end{table}

\begin{figure}[!h]
\vspace{-0.2in}
\centering
\begin{subfigure}{.5\textwidth}
  \centering
  \includegraphics[width=0.8\textwidth]{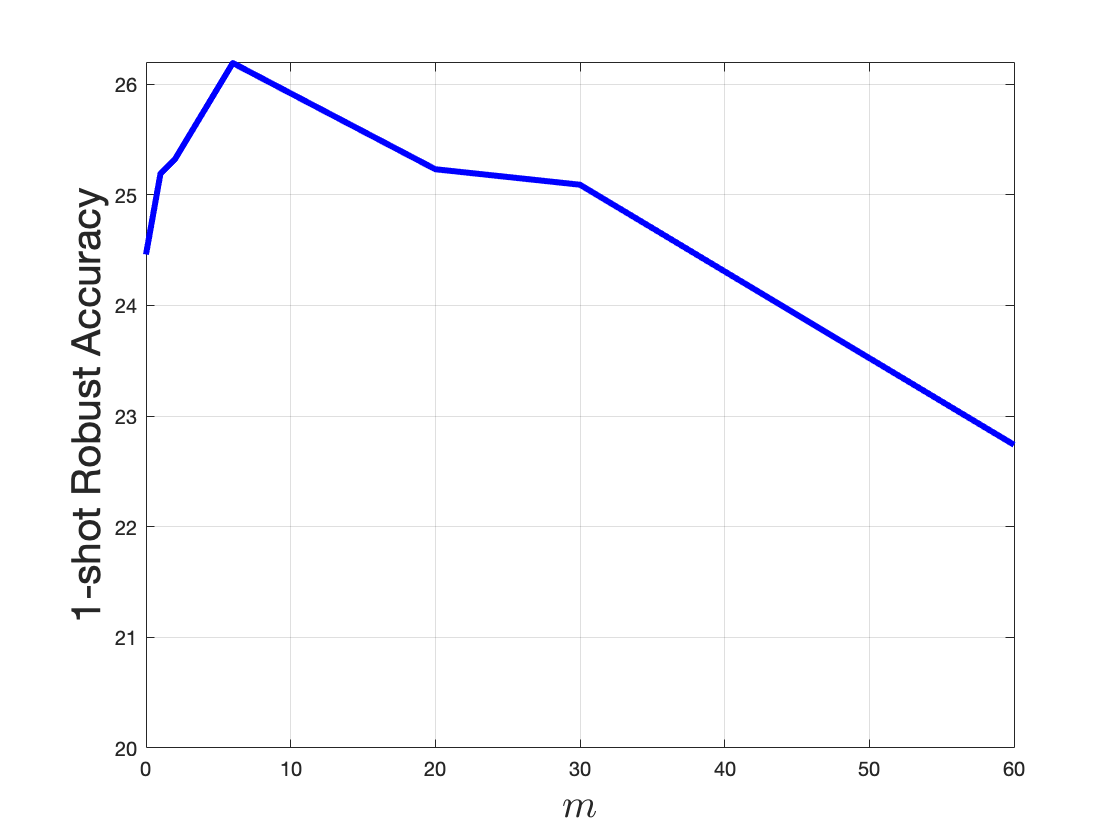}
  \caption{1-shot}
  \label{fig:sub1}
\end{subfigure}%
\begin{subfigure}{.5\textwidth}
  \centering
  \includegraphics[width=0.8\textwidth]{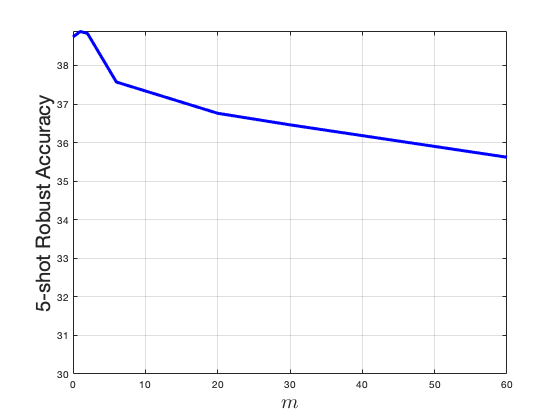}
  \caption{5-shot}
  \label{fig:sub2}
\end{subfigure}
\caption{Variation of Robust accuracy with number of base centers $m$ for 1-shot and 5-shot settings}
\vspace{-0.4in}
\label{fig:var_m_rob}
\end{figure}

 {\noindent \bf Varying the number of base categories chosen:} We plot the variation of mean Robust Accuracy with number of base neighbors $m$ in Figure \ref{fig:var_m_rob}.
 When $m=0$, the method becomes similar to a Nearest Centroid Classifier without calibration. We find best results with $m=2$ which we use for all our experiments. This means that centroids from $m=2$ base categories is sufficient to boost the performance of our model. Note that this calibration step is performed once prior to the inference and can be considered a preprocessing step, hence not affecting the inference time. This experiment also allows to study the relationship between the base and novel datasets, which although have no overlap in categories share some similarity in terms of transferable abstract level concepts. Such a setting is difficult to consider for episodic baselines, since they observe only few categories per task and hence do not contain information across multiple categories.  We would like to emphasize that we use the same dataset as previous meta-learning based approaches, only that our base training is standard mini-batch training.

  \begin{table}[!h]
    \centering
   \begin{tabular}{|l ||c| c| }
\hline
 &  \multicolumn{2}{|c|}{1-shot } \\
\hline
  Method &   Standard & Robust\\
 &   Acc.  & Acc.\\
\hline

Linear No Adv & 42.63  $\pm$ 0.56   & 19.56 $\pm$ 0.45    \\
Linear 7-PGD & 42.00  $\pm$  0.56   & 18.83 $\pm$ 0.42    \\
Linear 20-PGD  & 42.03   $\pm$  0.58   & 19.01  $\pm$ 0.42     \\
\hline
Ours (CNC) & \textbf{44.98  $\pm$ 0.59 }  & \textbf{21.38 $\pm$ 0.46 }   \\

\hline

\hline
\end{tabular}

    \caption{Experiment where adversarial examples from few-shot data are included to perform adversarial training on the linear classifier.  \textbf{No Adv} refers to clean examples used for learning the linear layer. \textbf{7-PGD} refers to 7-step PGD and \textbf{20-PGD} refers to 20-step PGD used in training. Robust Accuracy is calculated using 20-step PGD. Our method outperforms on both metrics without increasing computation that arises due to adversarial training.  }
    \label{tab:adv_test}
    \vspace{-0.4in}
\end{table}

{ \noindent \bf Using Adversarial examples from few-shot categories:} We additionally consider an experiment where adversarial examples of the few-shot data are used when learning the linear classifier. Although this makes the learning more complex due to the adversarial training procedure, we perform this baseline experiment and learn the linear layer.We use a 7-step PGD and 20-step PGD when learning the classifier. As seen in Table \ref{tab:adv_test} we observe a drop in performance which suggests that the network might be biased towards adversary of the few-shot data. This suggests that creating adversarial examples increases computation but does not lead to performance gains. Our CNC method outperforms previous methods without increasing computational burden due to creation of adversarial examples. This experiment uses ResNet18 backbone on Mini-ImageNet dataset.

\newpage
\section{Conclusion}
We present a simple and scalable approach for improving robustness in few-shot image classifiers. Our method outperforms previous approaches when compared with standard few-shot learning benchmarks on both standard and robust accuracy. Note that our method is similar to traditional adversarial machine learning approaches rather than meta-learning methods, hence it becomes easier to introduce concepts such as certified robustness which is unexplored for few-shot classifiers. We show preliminary results and plan to explore this direction in future works. We believe that the simplicity of our approach would be beneficial to the community, upon which researchers can develop robust few-shot classifiers.

{\noindent \bf Acknowledgment:}
This material is based upon work partially supported by the United States Air Force under Contract No. FA8750‐19‐C‐0098, funding from SAP SE, NSF grant 1845216, and also financial assistance award number 60NANB18D279 from U.S. Department of Commerce, National Institute of Standards and Technology. Any opinions, findings, and conclusions or recommendations expressed in this material are those of the authors and do not necessarily reflect the views of the United States Air Force, DARPA, or other funding agencies.

\bibliographystyle{splncs04}
\bibliography{egbib}

\begin{thebibliography}{10}
\providecommand{\url}[1]{\texttt{#1}}
\providecommand{\urlprefix}{URL }
\providecommand{\doi}[1]{https://doi.org/#1}

\bibitem{andrychowicz2016learning}
Andrychowicz, M., Denil, M., Gomez, S., Hoffman, M.W., Pfau, D., Schaul, T.,
  Shillingford, B., De~Freitas, N.: Learning to learn by gradient descent by
  gradient descent. arXiv preprint arXiv:1606.04474  (2016)

\bibitem{antoniou2017data}
Antoniou, A., Storkey, A., Edwards, H.: Data augmentation generative
  adversarial networks. arXiv preprint arXiv:1711.04340  (2017)

\bibitem{athalye2018obfuscated}
Athalye, A., Carlini, N., Wagner, D.: Obfuscated gradients give a false sense
  of security: Circumventing defenses to adversarial examples. In:
  International Conference on Machine Learning. pp. 274--283 (2018)

\bibitem{bertinetto2016learning}
Bertinetto, L., Henriques, J.F., Valmadre, J., Torr, P.H., Vedaldi, A.:
  Learning feed-forward one-shot learners. arXiv preprint arXiv:1606.05233
  (2016)

\bibitem{bishop:2006:PRML}
Bishop, C.M.: Pattern Recognition and Machine Learning. Springer (2006)

\bibitem{carlini2016defensive}
Carlini, N., Wagner, D.: Defensive distillation is not robust to adversarial
  examples

\bibitem{carlini2017adversarial}
Carlini, N., Wagner, D.: Adversarial examples are not easily detected:
  Bypassing ten detection methods. In: Proceedings of the 10th ACM Workshop on
  Artificial Intelligence and Security. pp. 3--14. ACM (2017)

\bibitem{chen2021robust}
Chen, T., Zhang, Z., Liu, S., Chang, S., Wang, Z.: Robust overfitting may be
  mitigated by properly learned smoothening

\bibitem{chen2020simple}
Chen, T., Kornblith, S., Norouzi, M., Hinton, G.: A simple framework for
  contrastive learning of visual representations. In: International conference
  on machine learning. pp. 1597--1607. PMLR (2020)

\bibitem{chen2019closer}
Chen, W.Y., Liu, Y.C., Kira, Z., Wang, Y.C.F., Huang, J.B.: A closer look at
  few-shot classification. arXiv preprint arXiv:1904.04232  (2019)

\bibitem{cohen2019certified}
Cohen, J., Rosenfeld, E., Kolter, Z.: Certified adversarial robustness via
  randomized smoothing. In: International Conference on Machine Learning. pp.
  1310--1320. PMLR (2019)

\bibitem{dhillon2019baseline}
Dhillon, G.S., Chaudhari, P., Ravichandran, A., Soatto, S.: A baseline for
  few-shot image classification. arXiv preprint arXiv:1909.02729  (2019)

\bibitem{feinman2017detecting}
Feinman, R., Curtin, R.R., Shintre, S., Gardner, A.B.: Detecting adversarial
  samples from artifacts. arXiv preprint arXiv:1703.00410  (2017)

\bibitem{finn2017model}
Finn, C., Abbeel, P., Levine, S.: Model-agnostic meta-learning for fast
  adaptation of deep networks. In: International Conference on Machine
  Learning. pp. 1126--1135. PMLR (2017)

\bibitem{garipov2018loss}
Garipov, T., Izmailov, P., Podoprikhin, D., Vetrov, D., Wilson, A.G.: Loss
  surfaces, mode connectivity, and fast ensembling of dnns. arXiv preprint
  arXiv:1802.10026  (2018)

\bibitem{gidaris2018dynamic}
Gidaris, S., Komodakis, N.: Dynamic few-shot visual learning without
  forgetting. In: Proceedings of the IEEE Conference on Computer Vision and
  Pattern Recognition. pp. 4367--4375 (2018)

\bibitem{goldblum2019robust}
Goldblum, M., Fowl, L., Goldstein, T.: Robust few-shot learning with
  adversarially queried meta-learners. arXiv preprint arXiv:1910.00982  (2019)

\bibitem{43405}
Goodfellow, I., Shlens, J., Szegedy, C.: Explaining and harnessing adversarial
  examples. In: International Conference on Learning Representations (2015),
  \url{http://arxiv.org/abs/1412.6572}

\bibitem{gowal2018effectiveness}
Gowal, S., Dvijotham, K., Stanforth, R., Bunel, R., Qin, C., Uesato, J.,
  Arandjelovic, R., Mann, T., Kohli, P.: On the effectiveness of interval bound
  propagation for training verifiably robust models. arXiv preprint
  arXiv:1810.12715  (2018)

\bibitem{gowal2020uncovering}
Gowal, S., Qin, C., Uesato, J., Mann, T., Kohli, P.: Uncovering the limits of
  adversarial training against norm-bounded adversarial examples. arXiv
  preprint arXiv:2010.03593  (2020)

\bibitem{grill2020bootstrap}
Grill, J.B., Strub, F., Altch{\'e}, F., Tallec, C., Richemond, P.H.,
  Buchatskaya, E., Doersch, C., Pires, B.A., Guo, Z.D., Azar, M.G., et~al.:
  Bootstrap your own latent: A new approach to self-supervised learning. arXiv
  preprint arXiv:2006.07733  (2020)

\bibitem{Guo_2020_cvpr}
Guo, J., Zhu, X., Zhao, C., Cao, D., Lei, Z., Li, S.Z.: Learning meta face
  recognition in unseen domains. In: Proceedings of the IEEE/CVF Conference on
  Computer Vision and Pattern Recognition (CVPR) (June 2020)

\bibitem{hariharan2017low}
Hariharan, B., Girshick, R.: Low-shot visual recognition by shrinking and
  hallucinating features. In: Proceedings of the IEEE International Conference
  on Computer Vision. pp. 3018--3027 (2017)

\bibitem{He_2016_CVPR}
He, K., Zhang, X., Ren, S., Sun, J.: Deep residual learning for image
  recognition. In: Proceedings of the IEEE Conference on Computer Vision and
  Pattern Recognition (CVPR) (June 2016)

\bibitem{hilliard2018few}
Hilliard, N., Phillips, L., Howland, S., Yankov, A., Corley, C.D., Hodas, N.O.:
  Few-shot learning with metric-agnostic conditional embeddings. arXiv preprint
  arXiv:1802.04376  (2018)

\bibitem{huang2017densely}
Huang, G., Liu, Z., Van Der~Maaten, L., Weinberger, K.Q.: Densely connected
  convolutional networks. In: Proceedings of the IEEE conference on computer
  vision and pattern recognition. pp. 4700--4708 (2017)

\bibitem{izmailov2018averaging}
Izmailov, P., Podoprikhin, D., Garipov, T., Vetrov, D., Wilson, A.G.: Averaging
  weights leads to wider optima and better generalization. arXiv preprint
  arXiv:1803.05407  (2018)

\bibitem{world-arxiv-2016}
Kurakin, A., Goodfellow, I., Bengio, S.: Adversarial examples in the physical
  world. arXiv preprint arXiv:1607.02533  (2016)

\bibitem{filterstats-arxiv-2016}
Li, X., Li, F.: Adversarial examples detection in deep networks with
  convolutional filter statistics. arXiv preprint arXiv:1612.07767  (2016)

\bibitem{madry2018towards}
Madry, A., Makelov, A., Schmidt, L., Tsipras, D., Vladu, A.: Towards deep
  learning models resistant to adversarial attacks. In: International
  Conference on Learning Representations (2018),
  \url{https://openreview.net/forum?id=rJzIBfZAb}

\bibitem{maicas2018training}
Maicas, G., Bradley, A.P., Nascimento, J.C., Reid, I., Carneiro, G.: Training
  medical image analysis systems like radiologists. In: International
  Conference on Medical Image Computing and Computer-Assisted Intervention. pp.
  546--554. Springer (2018)

\bibitem{mirman2018differentiable}
Mirman, M., Gehr, T., Vechev, M.: Differentiable abstract interpretation for
  provably robust neural networks. In: International Conference on Machine
  Learning. pp. 3578--3586. PMLR (2018)

\bibitem{nichol2018first}
Nichol, A., Achiam, J., Schulman, J.: On first-order meta-learning algorithms.
  arXiv preprint arXiv:1803.02999  (2018)

\bibitem{papernot2016distillation}
Papernot, N., McDaniel, P., Wu, X., Jha, S., Swami, A.: Distillation as a
  defense to adversarial perturbations against deep neural networks. In:
  Security and Privacy (SP), 2016 IEEE Symposium on. pp. 582--597. IEEE (2016)

\bibitem{paszke2017automatic}
Paszke, A., Gross, S., Chintala, S., Chanan, G., Yang, E., DeVito, Z., Lin, Z.,
  Desmaison, A., Antiga, L., Lerer, A.: Automatic differentiation in pytorch
  (2017)

\bibitem{qi2018low}
Qi, H., Brown, M., Lowe, D.G.: Low-shot learning with imprinted weights. In:
  Proceedings of the IEEE conference on computer vision and pattern
  recognition. pp. 5822--5830 (2018)

\bibitem{raghunathan2018semidefinite}
Raghunathan, A., Steinhardt, J., Liang, P.: Semidefinite relaxations for
  certifying robustness to adversarial examples. arXiv preprint
  arXiv:1811.01057  (2018)

\bibitem{rakelly2018conditional}
Rakelly, K., Shelhamer, E., Darrell, T., Efros, A., Levine, S.: Conditional
  networks for few-shot semantic segmentation  (2018)

\bibitem{ravi2016optimization}
Ravi, S., Larochelle, H.: Optimization as a model for few-shot learning  (2016)

\bibitem{ravichandran2019few}
Ravichandran, A., Bhotika, R., Soatto, S.: Few-shot learning with embedded
  class models and shot-free meta training. In: Proceedings of the IEEE/CVF
  International Conference on Computer Vision. pp. 331--339 (2019)

\bibitem{ren2018meta}
Ren, M., Triantafillou, E., Ravi, S., Snell, J., Swersky, K., Tenenbaum, J.B.,
  Larochelle, H., Zemel, R.S.: Meta-learning for semi-supervised few-shot
  classification. arXiv preprint arXiv:1803.00676  (2018)

\bibitem{ILSVRC15}
Russakovsky, O., Deng, J., Su, H., Krause, J., Satheesh, S., Ma, S., Huang, Z.,
  Karpathy, A., Khosla, A., Bernstein, M., Berg, A.C., Fei-Fei, L.: {ImageNet
  Large Scale Visual Recognition Challenge}. International Journal of Computer
  Vision (IJCV)  (2015). \doi{10.1007/s11263-015-0816-y}

\bibitem{schmidt2018adversarially}
Schmidt, L., Santurkar, S., Tsipras, D., Talwar, K., M{\k{a}}dry, A.:
  Adversarially robust generalization requires more data. arXiv preprint
  arXiv:1804.11285  (2018)

\bibitem{shafahi2019adversarial}
Shafahi, A., Najibi, M., Ghiasi, A., Xu, Z., Dickerson, J., Studer, C., Davis,
  L.S., Taylor, G., Goldstein, T.: Adversarial training for free! arXiv
  preprint arXiv:1904.12843  (2019)

\bibitem{snell2017prototypical}
Snell, J., Swersky, K., Zemel, R.S.: Prototypical networks for few-shot
  learning. arXiv preprint arXiv:1703.05175  (2017)

\bibitem{sung2018learning}
Sung, F., Yang, Y., Zhang, L., Xiang, T., Torr, P.H., Hospedales, T.M.:
  Learning to compare: Relation network for few-shot learning. In: Proceedings
  of the IEEE conference on computer vision and pattern recognition. pp.
  1199--1208 (2018)

\bibitem{intriguing-arxiv-2013}
Szegedy, C., Zaremba, W., Sutskever, I., Bruna, J., Erhan, D., Goodfellow,
  I.J., Fergus, R.: Intriguing properties of neural networks. CoRR
  \textbf{abs/1312.6199} (2013), \url{http://arxiv.org/abs/1312.6199}

\bibitem{tsipras2018robustness}
Tsipras, D., Santurkar, S., Engstrom, L., Turner, A., Madry, A.: Robustness may
  be at odds with accuracy. arXiv preprint arXiv:1805.12152  (2018)

\bibitem{tukey1977exploratory}
Tukey, J.W., et~al.: Exploratory data analysis, vol.~2. Reading, Mass. (1977)

\bibitem{vinyals2016matching}
Vinyals, O., Blundell, C., Lillicrap, T., Kavukcuoglu, K., Wierstra, D.:
  Matching networks for one shot learning. arXiv preprint arXiv:1606.04080
  (2016)

\bibitem{wah2011caltech}
Wah, C., Branson, S., Welinder, P., Perona, P., Belongie, S.: The caltech-ucsd
  birds-200-2011 dataset  (2011)

\bibitem{wang2020tracking}
Wang, G., Luo, C., Sun, X., Xiong, Z., Zeng, W.: Tracking by instance
  detection: A meta-learning approach. In: Proceedings of the IEEE/CVF
  Conference on Computer Vision and Pattern Recognition. pp. 6288--6297 (2020)

\bibitem{wang2021fast}
Wang, R., Xu, K., Liu, S., Chen, P.Y., Weng, T.W., Gan, C., Wang, M.: On fast
  adversarial robustness adaptation in model-agnostic meta-learning. arXiv
  preprint arXiv:2102.10454  (2021)

\bibitem{wang2018low}
Wang, Y.X., Girshick, R., Hebert, M., Hariharan, B.: Low-shot learning from
  imaginary data. In: Proceedings of the IEEE conference on computer vision and
  pattern recognition. pp. 7278--7286 (2018)

\bibitem{wong2018provable}
Wong, E., Kolter, Z.: Provable defenses against adversarial examples via the
  convex outer adversarial polytope. In: International Conference on Machine
  Learning. pp. 5286--5295. PMLR (2018)

\bibitem{wong2018scaling}
Wong, E., Schmidt, F.R., Metzen, J.H., Kolter, J.Z.: Scaling provable
  adversarial defenses. arXiv preprint arXiv:1805.12514  (2018)

\bibitem{xie2017adversarial}
Xie, C., Wang, J., Zhang, Z., Zhou, Y., Xie, L., Yuille, A.L.: Adversarial
  examples for semantic segmentation and object detection. CoRR
  \textbf{abs/1703.08603} (2017), \url{http://arxiv.org/abs/1703.08603}

\bibitem{xu2020automatic}
Xu, K., Shi, Z., Zhang, H., Wang, Y., Chang, K.W., Huang, M., Kailkhura, B.,
  Lin, X., Hsieh, C.J.: Automatic perturbation analysis for scalable certified
  robustness and beyond. Advances in Neural Information Processing Systems
  \textbf{33} (2020)

\bibitem{yang2020randomized}
Yang, G., Duan, T., Hu, J.E., Salman, H., Razenshteyn, I., Li, J.: Randomized
  smoothing of all shapes and sizes. In: International Conference on Machine
  Learning. pp. 10693--10705. PMLR (2020)

\bibitem{yang2021free}
Yang, S., Liu, L., Xu, M.: Free lunch for few-shot learning: Distribution
  calibration. In: International Conference on Learning Representations (ICLR)
  (2021)

\bibitem{yin2018adversarial}
Yin, C., Tang, J., Xu, Z., Wang, Y.: Adversarial meta-learning. arXiv preprint
  arXiv:1806.03316  (2018)

\bibitem{zagoruyko2016wide}
Zagoruyko, S., Komodakis, N.: Wide residual networks. arXiv preprint
  arXiv:1605.07146  (2016)

\bibitem{zhang2019theoretically}
Zhang, H., Yu, Y., Jiao, J., Xing, E., El~Ghaoui, L., Jordan, M.: Theoretically
  principled trade-off between robustness and accuracy. In: International
  Conference on Machine Learning. pp. 7472--7482. PMLR (2019)

\bibitem{zhang2019towards}
Zhang, H., Chen, H., Xiao, C., Gowal, S., Stanforth, R., Li, B., Boning, D.,
  Hsieh, C.J.: Towards stable and efficient training of verifiably robust
  neural networks. arXiv preprint arXiv:1906.06316  (2019)

\end{thebibliography}
\newpage
\appendix
\section{Appendix}

\noindent{\bf Variation of Robust Accuracy with number of attack iterations:} We vary the number of attack iterations of PGD and observe a fairly stable performance for both 1-shot and 5-shot settings, as seen in Table \ref{tab:attack_iter}. This experiment shows that defense is not sensitive to the number of attack iterations.

 \begin{table}[!h]
    \centering
    \resizebox{0.4\columnwidth}{!}{
   \begin{tabular}{|l ||c|c |  }
\hline
 &  \multicolumn{1}{|c|}{1-shot } & \multicolumn{1}{|c|}{5-shot }\\
\hline
PGD &   Robust & Robust\\
Iterations &   Acc.  & Acc.\\
\hline

20 & 25.32  $\pm$ 0.52   & 38.83 $\pm$ 0.57    \\
40 & 25.19  $\pm$ 0.53   & 38.46 $\pm$ 0.54    \\
100 & 25.64  $\pm$ 0.53   & 38.22 $\pm$ 0.56    \\
200 & 25.09  $\pm$ 0.54   & 38.62 $\pm$ 0.57    \\

\hline

\hline

\end{tabular}}
    \caption{Variation of attack iterations}
    \label{tab:attack_iter}
\end{table}

\begin{figure}[!h]
\centering
\begin{subfigure}{.5\textwidth}
  \centering
  \includegraphics[width=0.7\textwidth]{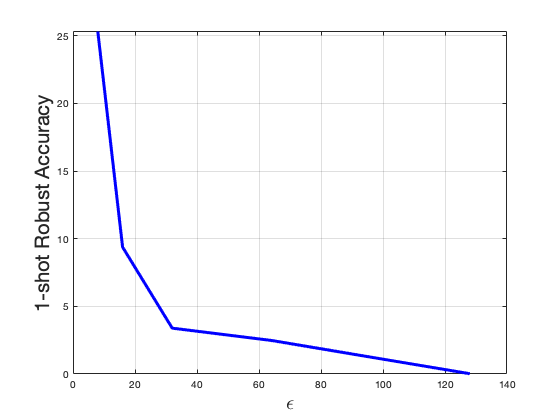}
  \caption{1-shot}
  \label{fig:sub1}
\end{subfigure}%
\begin{subfigure}{.5\textwidth}
  \centering
  \includegraphics[width=0.7\textwidth]{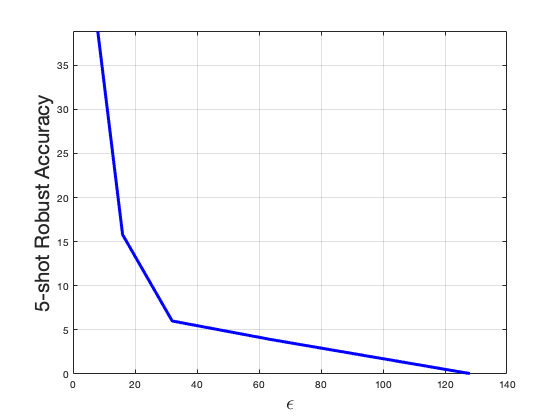}
  \caption{5-shot}
  \label{fig:sub2}
\end{subfigure}
\caption{Variation of Robust Accuracy with different perturbation budget $\epsilon$}
\label{fig:var_eps_m_rob}
\end{figure}

\noindent{\bf Variation of Robust Accuracy with perturbation budget $\epsilon$:} To check for the absence of gradient masking, we increase $\epsilon$ from 8/255 to 128/255 in Figure \ref{fig:var_eps_m_rob}. As expected, we observe that both 1-shot and 5-shot accuracy drop to zero with increased $\epsilon$. Note that we plot only the the mean accuracy over 1000 different tasks.

\begin{figure}[!h]
\centering
\begin{subfigure}{.5\textwidth}
  \centering
  \includegraphics[width=0.8\textwidth]{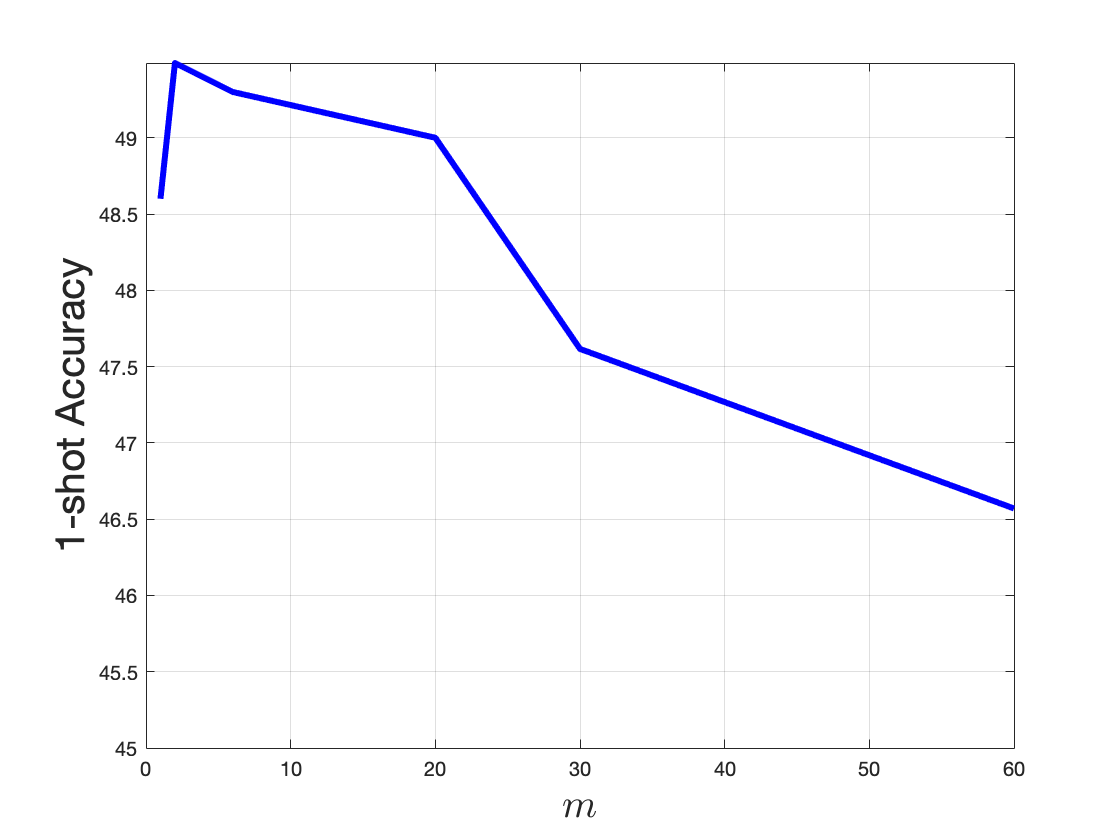}
  \caption{1-shot}
  \label{fig:sub1}
\end{subfigure}%
\begin{subfigure}{.5\textwidth}
  \centering
  \includegraphics[width=0.8\textwidth]{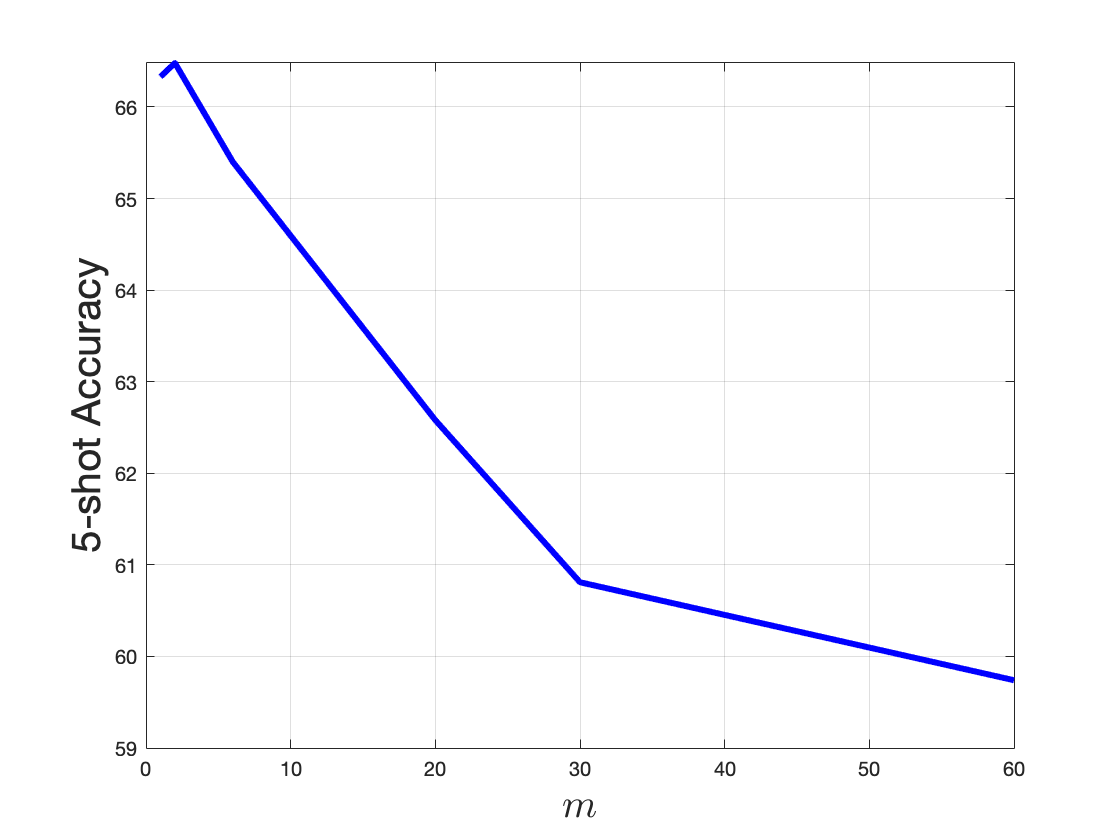}
  \caption{5-shot}
  \label{fig:sub2}
\end{subfigure}
\caption{Variation of Standard accuracy with number of base centers }
\label{fig:var_m}
\end{figure}

\noindent{\bf Number of base categories chosen:} We plot the variation of standard accuracy with $m$ in Figure \ref{fig:var_m} for 1-shot and 5-shot settings. It is to be noted that we plot only the mean accuracy averaged over 1000 test episodes. We find best results with $m=2$ which we use for all our experiments. \\

\section{Societal Impact}
We believe most AI algorithms including ours can be exploited by adversaries for unethical applications. On the positive side, we are developing algorithms that make deep networks robust to adversarial attacks, eliminating some of such undesired applications. Moreover, while publishing such defense algorithms may enable a larger community to access and benefit from them, leading to democratized AI, it can also enable adversaries to develop better attacks.

To ensure that our results are reproducible, we include all implementation details of our work and also provide code. We believe that readers will have sufficient knowledge of the hyperparameters associated with our method to reproduce the results.


\end{document}